\documentclass[acmtog,nonacm]{acmart}

\usepackage{booktabs}
\usepackage{graphicx}
\usepackage{amsmath}

\usepackage{amssymb}
\usepackage{multirow}
\usepackage{xcolor}
\usepackage{makecell}
\usepackage{colortbl}
\usepackage{pifont}
\usepackage{threeparttable}
\usepackage{placeins}
\usepackage{wasysym}
\usepackage{longtable}
\usepackage{array}
\usepackage{float}
\usepackage{hyperref}
\definecolor{linkblue}{HTML}{4E8EF7}
\hypersetup{
    colorlinks=true,
    urlcolor=linkblue,
    linkcolor=black,
    citecolor=black
}
\usepackage[ruled]{algorithm2e}

\SetAlFnt{\small}
\SetAlCapFnt{\small}
\SetAlCapNameFnt{\small}
\SetAlCapHSkip{0pt}

\acmJournal{TOG}

\newcommand{\ours}{Tipiano}

\begin{document}

\title{Tipiano: Cascaded Piano Hand Motion Synthesis via Fingertip Priors}

\author{Joonhyung Bae}
\orcid{0000-0001-5933-4302}
\affiliation{%
  \institution{Korea Advanced Institute of Science and Technology (KAIST)}
  \city{Daejeon}
  \country{South Korea}}
\email{jh.bae@kaist.ac.kr}

\author{Kirak Kim}
\orcid{0000-0002-2960-4583}
\affiliation{%
  \institution{Korea Advanced Institute of Science and Technology (KAIST)}
  \city{Daejeon}
  \country{South Korea}}
\email{kirak@kaist.ac.kr}

\author{Hyeyoon Cho}
\orcid{0000-0003-2926-941X}
\affiliation{%
  \institution{Korea Advanced Institute of Science and Technology (KAIST)}
  \city{Daejeon}
  \country{South Korea}}
\email{hyeyooncho@kaist.ac.kr}

\author{Sein Lee}
\orcid{0009-0005-0746-8065}
\affiliation{%
  \institution{Korea Advanced Institute of Science and Technology (KAIST)}
  \city{Daejeon}
  \country{South Korea}}
\email{seinlee@kaist.ac.kr}

\author{Yoon-Seok Choi}
\orcid{0009-0003-0950-3253}
\affiliation{%
  \institution{Seoul National University}
  \city{Seoul}
  \country{South Korea}}
\email{dallas71@snu.ac.kr}

\author{Hyeon Hur}
\orcid{0009-0008-8662-2121}
\affiliation{%
  \institution{Seoul National University}
  \city{Seoul}
  \country{South Korea}}
\email{sunny000927@snu.ac.kr}

\author{Gyubin Lee}
\orcid{0009-0007-5274-051X}
\affiliation{%
  \institution{Korea Advanced Institute of Science and Technology (KAIST)}
  \city{Daejeon}
  \country{South Korea}}
\email{gbstorm81@kaist.ac.kr}

\author{Akira Maezawa}
\affiliation{%
  \institution{YAMAHA}
  \city{Hamamatsu}
  \country{Japan}}
\email{akira.maezawa@music.yamaha.com}

\author{Satoshi Obata}
\affiliation{%
  \institution{YAMAHA}
  \city{Hamamatsu}
  \country{Japan}}
\email{satoshi.obata@music.yamaha.com}

\author{Jonghwa Park}
\affiliation{%
  \institution{Seoul National University}
  \city{Seoul}
  \country{South Korea}}
\email{pianistpark@gmail.com}

\author{Jaebum Park}
\affiliation{%
  \institution{Seoul National University}
  \city{Seoul}
  \country{South Korea}}
\email{parkpe95@snu.ac.kr}

\author{Juhan Nam}
\authornote{Corresponding author.}
\orcid{0000-0003-2664-2119}
\affiliation{%
  \institution{Korea Advanced Institute of Science and Technology (KAIST)}
  \city{Daejeon}
  \country{South Korea}}
\email{juhan.nam@kaist.ac.kr}

\begin{teaserfigure}
\centering
\includegraphics[width=1.0\linewidth]{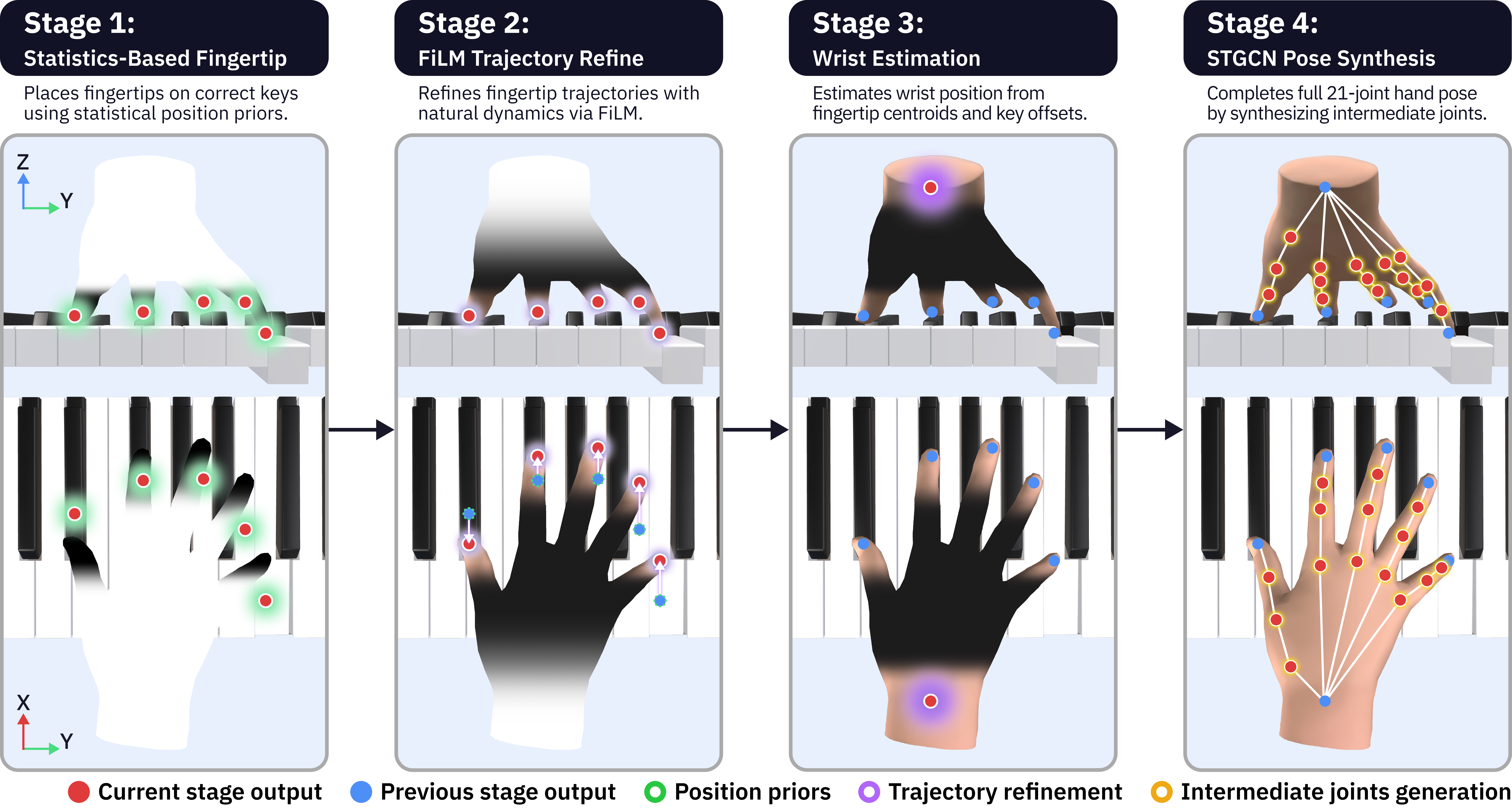}
\vspace{-6pt}
\caption{Overview of \ours{}. Our four-stage cascade exploits decreasing ambiguity from fingertips to intermediate joints.}
\label{fig:teaser}
\end{teaserfigure}

\begin{abstract}
Synthesizing realistic piano hand motions requires both precision and naturalness.
Physics-based methods achieve precision but produce stiff motions; data-driven models learn natural dynamics but struggle with positional accuracy.
Piano motion exhibits a natural hierarchy: fingertip positions are nearly deterministic given piano geometry and fingering, while wrist and intermediate joints offer stylistic freedom.
We present \ours{}, a four-stage framework exploiting this hierarchy: (1) statistics-based fingertip positioning, (2) FiLM-conditioned trajectory refinement, (3) wrist estimation, and (4) STGCN-based pose synthesis.
We contribute expert-annotated fingerings for the FürElise dataset (153 pieces, ${\sim}$10 hours).
Experiments demonstrate F1 = 0.910, substantially outperforming diffusion baselines (F1 = 0.121), with user study ($N=41$) confirming quality approaching motion capture.
Expert evaluation by professional pianists ($N=5$) identified anticipatory motion as the key remaining gap, providing concrete directions for future improvement.
\end{abstract}



\maketitle

\vspace{-3pt}
\section{Introduction}

Piano playing demands precision, coordination, and dexterity~\cite{10.1145/3680528.3687703}.
Synthesizing realistic piano hand motions enables applications in character animation, embodied AI, music education, and VR/AR.
Unlike general motion synthesis, piano performance is \textit{functionally constrained}---motion must produce specific notes at specific times.

Two paradigms address this challenge with distinct trade-offs.
\textbf{Physics-based RL}~\cite{10.1145/3680528.3687703,robopianist2023} trains policies that physically interact with piano keys, achieving precision but requiring expensive per-piece training (1--3 days) and producing mechanically stiff motions.
\textbf{Data-driven generative models}~\cite{gan2024pianomotion,10.1145/3746027.3755097} learn from motion capture, producing natural dynamics but struggling with positional accuracy---fingers move expressively yet miss target keys.

This trade-off stems from different sources in each paradigm:
physics-based RL optimizes for task success without naturalness objectives, while data-driven models employ \textit{monolithic} generation 
conflating geometric constraints with stylistic variation.
Our key insight is that piano motion exhibits a \textit{hierarchy of constraints}:
(1) \textbf{Fingertip positions} are constrained---fingertips \textit{must} contact the target key's surface;
(2) \textbf{Wrist trajectories} must position fingers within reach but allow stylistic variation;
(3) \textbf{Intermediate joints} are kinematically constrained by endpoints but retain curling freedom.

This hierarchy suggests solving the most constrained sub-problem first, then progressively adding freedom while respecting prior constraints.
Enforcing geometric constraints early avoids post-hoc correction, which degrades learned dynamics or introduces artifacts.

We present \ours{}, implementing this decomposition across four cascaded stages (Figure~\ref{fig:teaser}): (1) statistics-based fingertip positioning, (2) FiLM-conditioned~\cite{perez2018film} trajectory refinement, (3) wrist estimation via key-based offsets, and (4) STGCN-based~\cite{stgcn} full-hand pose synthesis.

To enable this approach, we contribute expert-annotated fingerings for the FürElise dataset~\cite{10.1145/3680528.3687703}, covering 153 pieces (340K notes, ${\sim}$10 hours) with per-frame finger-to-key assignments that directly yield position priors.

\noindent\textbf{Contributions:}
\begin{itemize}
    \item A four-stage hierarchical framework combining deterministic priors with learned refinement.
    \item Expert fingering annotations for the FürElise dataset: 340K per-frame finger-key contact assignments with gesture boundaries, enabling the position prior construction central to our framework.
    \item Empirical validation achieving F1 = 0.910 for key contact accuracy---substantially outperforming diffusion baselines (F1 = 0.121)---while maintaining naturalness comparable to the FürElise dataset.
\end{itemize}
\section{Related Work}

\subsection{Music-Driven Motion Synthesis}
Music-conditioned dance generation has achieved compelling results using diffusion models and Transformer architectures~\cite{edge,li2021ai,alexanderson2023listen}.
Audio-driven synthesis~\cite{shlizerman2018audio,li2018skeleton} struggles with fine-grained hand articulation.
These approaches prioritize aesthetic plausibility; piano performance additionally requires positional accuracy for correct key contact.

\subsection{Piano Hand Motion Synthesis}

Prior work falls into physics-based and data-driven paradigms, reflecting the precision-naturalness trade-off.

\textbf{Early approaches} combined rule-based modeling with motion capture priors.
Yamamoto et al.~\cite{5650193} constructed a position prior from motion capture data and generated hand poses using inverse kinematics, though without learned refinement.
Zhu et al.~\cite{zhu2013system} proposed a fingering generation algorithm based on geometric constraints and graph-based optimization.
While ensuring geometric validity, these methods relied on handcrafted rules.

\textbf{Physics-based approaches} formulate piano performance as a control problem, training RL policies on simulated pianos.
RoboPianist~\cite{robopianist2023} achieves functional key pressing but produces robot-like motion.
FürElise~\cite{10.1145/3680528.3687703} combines diffusion with RL, achieving precision but requiring expensive per-piece training (1--3 days) with stiff outputs.
Recent work~\cite{qian2024pianomimelearninggeneralistdexterous,huang2025pandora} demonstrates dexterous control but prioritizes task success over expressiveness.

\textbf{Data-driven approaches} learn hand motion from motion capture without physical simulation.
Early efforts~\cite{li2018skeleton} trained LSTM generators but struggled with fine-grained hand articulation.
PianoMotion10M~\cite{gan2024pianomotion} provides large-scale data (116 hours) with diffusion baselines.
S2C~\cite{10.1145/3746027.3755097} introduces dual-stream diffusion with Hand-Coordinated Asymmetric Attention for bimanual coordination. These methods produce natural motion but exhibit inaccurate key contacts.

\subsection{Piano Performance Datasets}
Dexterous hand control has broad applications~\cite{andrychowicz2020learning,zhang2021manipnet,yang2022learning,mordatch2012contact}, with datasets for grasping~\cite{taheri2020grab}, manipulation~\cite{fan2023arctic}, and two-hand interaction~\cite{moon2020interhand2}, though most lack piano's precision requirements.

Prior datasets~\cite{simon2017hand,wu2023marker} were limited in scale or lacked synchronized audio; FürElise~\cite{10.1145/3680528.3687703} provides high-quality markerless capture with MIDI synchronization, enabling our position prior analysis.
Piano fingering---assigning fingers to keys---determines biomechanical positioning constraints.
Table~\ref{tab:fingering_datasets} compares existing datasets.
PIG~\cite{NAKAMURA202068} provides expert annotations (100K notes) but only score-based annotations without performance data.
ThumbSet~\cite{10.1145/3503161.3548372} offers larger scale but relies on crowdsourced amateur annotations.
PianoVAM~\cite{kim2025pianovam} provides multimodal data but captures amateur practice rather than professional performance.
Unlike hand-object datasets providing contact areas, piano fingering offers \textit{per-frame finger-to-key assignments} where target positions are deterministic given piano geometry.

\begin{table}[t]
\centering
\vspace{-6pt}
\caption{Piano fingering datasets comparison. Fingering provides per-note finger-key contact assignments, enabling position prior construction.}
\label{tab:fingering_datasets}
\vspace{-3pt}
\resizebox{\columnwidth}{!}{
\begin{threeparttable}
\begin{tabular}{lcccccc}
\toprule
\textbf{Dataset} & \textbf{Pieces} & \textbf{Notes} & \textbf{Hours} & \textbf{Performer} & \textbf{Annotation} & \textbf{Boundary} \\
\midrule
PIG$^{a}$ & 150 & 100K & — & Expert & Expert & \ding{55} \\
ThumbSet$^{b}$ & 2.5K & — & — & — & Amateur & \ding{55} \\
PianoVAM$^{c}$ & 106 & 1.05M & 21 & Amateur & Researcher & \ding{55} \\
\textbf{\ours{}} & \textbf{153} & \textbf{340K} & \textbf{10} & \textbf{Expert} & \textbf{Expert} & \ding{51} \\
\bottomrule
\end{tabular}
\begin{tablenotes}
\footnotesize
\item[a] \cite{NAKAMURA202068}, [b] \cite{10.1145/3503161.3548372}, [c] \cite{kim2025pianovam}. Boundary: gesture boundary.
\end{tablenotes}
\end{threeparttable}
\vspace{-8pt}
}
\end{table}

\section{The \ours{} Fingering Dataset}
\label{sec:dataset}

Prior piano datasets provide either fingerings without motion~\cite{NAKAMURA202068} or motion without reliable fingerings~\cite{gan2024pianomotion,10.1145/3680528.3687703}.
We introduce the \ours{} Dataset built upon FürElise~\cite{10.1145/3680528.3687703}, collected using a Yamaha Disklavier DS7X ENPRO: 153 pieces by 15 professional pianists (8 male, 7 female), comprising ${\sim}$10 hours of markerless motion capture at 59.94fps with synchronized audio and MIDI across 7 musical periods and 8 difficulty levels.

\textbf{Fingering as Contact Supervision.}
Each annotation specifies fingertip-key contacts per frame---yielding 340K events.
Annotations were produced through rule-based extraction (${\sim}$85\% accuracy) and expert review.

\textbf{Gesture Boundaries.}
Beyond per-note fingerings, we provide \textit{gesture boundary annotations} marking temporal segments where hands are placed on or lifted from the keyboard.
These boundaries delineate performance phrases, critical for synthesis at phrase boundaries.
To our knowledge, \ours{} is the first piano dataset with such segmentation.

\textbf{Metadata and Split.}
The original FürElise dataset lacks difficulty and period labels as well as train/val/test splits.
We contribute metadata annotations---difficulty levels (Henle scale 1--9) and musical periods---labeled by professional pianists, and introduce stratified splitting by difficulty (primary) and period (secondary), yielding 96/28/27 pieces for train/val/test.
Each split covers all 7 difficulty levels and 7 musical periods.

\textbf{Data Quality.}
Markerless capture inherently introduces fingertip positioning errors---FürElise reports F1 = 86.49\% for key contact detection~\cite{10.1145/3680528.3687703}.
Under our stricter evaluation threshold calibrated to the Yamaha Disklavier DS7X ENPRO's 8mm acoustic trigger depth---confirmed through direct consultation with Yamaha Corporation---this yields F1 = 0.739.
Qualitative assessment by an external piano expert and seven annotators characterized how these inaccuracies manifest: errors appear as vertical displacement (hovering above keys), horizontal displacement (positioning between keys), or temporal misalignment, concentrated in longer sequences and technically demanding passages.
Despite these limitations, the expert confirmed the dataset captures ``natural, efficient movement following gravity optimally''---professional technique difficult to obtain through marker-based approaches, where physical markers can hinder natural finger movements during performance, or simulated approaches.

\section{Method}
\label{sec:method}

\begin{figure*}[t]
\centering
\vspace{-4pt}
\includegraphics[width=\textwidth]{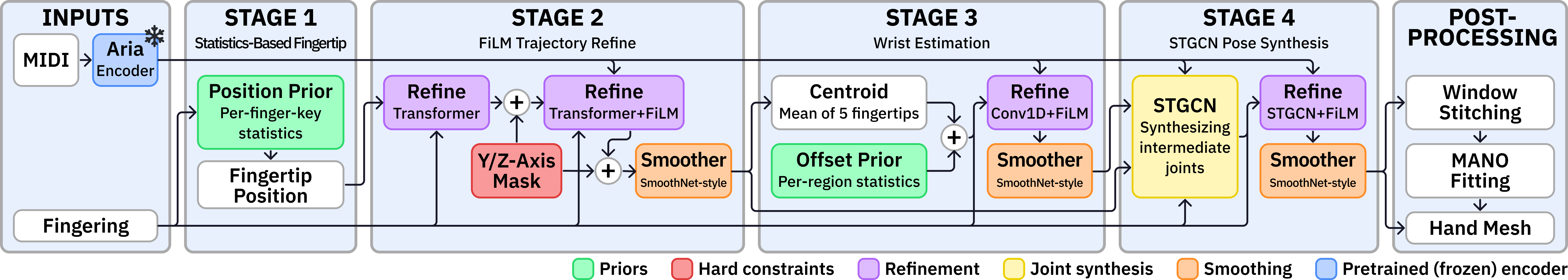}
\vspace{-8pt}
\caption{The \ours{} pipeline. From MIDI and fingering, four cascaded stages synthesize hand motion: (1) statistics-based fingertip positioning, (2) FiLM-conditioned trajectory refinement with geometric constraints, (3) wrist estimation via offset priors, and (4) STGCN-based full pose synthesis.}
\label{fig:architecture}
\end{figure*}

\subsection{Overview and Problem Formulation}
\label{sec:overview}

Our framework implements the hierarchical decomposition introduced in Section~1 through four cascaded stages (Figure~\ref{fig:architecture}).
We process 8-second windows (480 frames at 59.94fps), matching S2C's~\cite{10.1145/3746027.3755097} window length for fair comparison. This duration balances musical phrase coverage with computational tractability. Overlapping windows with 4-second stride are used for longer pieces (Section~\ref{sec:postprocessing}).

Given MIDI events $\mathbf{M} = \{(t_i, p_i, v_i, d_i)\}$ encoding onset, pitch, velocity, and duration of pressed keys, along with fingering annotations $\mathbf{F} \in \{0, 1, \ldots, 10\}^{T \times 88}$ specifying which finger presses each of 88 keys per frame, we synthesize hand joint positions $\mathbf{J}^{h} \in \mathbb{R}^{T \times 21 \times 3}$ for each hand $h \in \{L, R\}$, using piano-aligned coordinates (X: front-back toward pianist, Y: along keyboard, Z: height).
Each hand comprises 21 joints following MANO convention~\cite{mano}: 1 wrist and 4 joints per finger (MCP, PIP, DIP, fingertip).
Fingering values 1--5 denote left hand (thumb to pinky); 6--10 denote right hand; 0 indicates no press.

\subsection{Hierarchical Motion Synthesis}
\label{sec:hierarchical}

\subsubsection{Stage 1: Statistics-Based Fingertip Positioning}
\label{sec:stage1}

Our analysis reveals that fingertip positions of fingers actively pressing keys are nearly deterministic.
When finger $f$ of hand $h$ presses key $k$, the fingertip must contact that key's surface, with variation only in the precise contact point within the key's bounds.

\textbf{Position Prior Construction.}
We construct a position prior $\mathcal{C}$ from ground-truth training data by collecting all frames where each (hand, finger, key) combination occurs during active presses.
For each combination $(h, f, k)$, we compute robust statistics in piano-aligned coordinates:
\begin{equation}
    \mathcal{C}(h, f, k) = \{\boldsymbol{\mu}_{hfk}, \boldsymbol{\sigma}_{hfk}, \boldsymbol{p}_{25}, \boldsymbol{p}_{50}, \boldsymbol{p}_{75}, n_{hfk}\}
\end{equation}
where $\boldsymbol{\mu} \in \mathbb{R}^3$ is the mean position, $\boldsymbol{\sigma}$ is standard deviation, $\boldsymbol{p}_{25/50/75}$ are quartiles, and $n$ is observation count.
Only training data is used for prior construction to prevent leakage.

Analysis reveals 735 unique combinations with $n \geq 10$, : $\sigma_x = 16.3$mm (front-back), $\sigma_y = 11.4$mm (along keyboard), $\sigma_z = 7.7$mm (height)---all well within the standard white key width of 23.5mm, confirming positions are well-constrained within key surfaces.

\textbf{Coverage and Interpolation.}
The complete prior spans 880 entries (2 hands $\times$ 5 fingers $\times$ 88 keys), of which 735 (84\%) have sufficient observations ($n \geq 10$).
For unobserved combinations, we interpolate from neighboring keys along the keyboard axis, leveraging piano geometry's spatial regularity.

\textbf{Noise Reduction.}
Aggregating statistics across $n \geq 10$ observations per combination empirically yields more stable position estimates than individual frames. However, we do not characterize the error distribution of the markerless capture system; errors may exhibit systematic patterns correlated with specific techniques or passage types.

\textbf{Baseline Generation.}
For each frame, we generate baseline fingertip positions deterministically:
(1) \textit{Pressing fingers} use median position $\boldsymbol{p}_{50}$ from the prior;
(2) \textit{Non-pressing fingers} interpolate using pressing fingers as anchors, with a hover height prior of 14mm and finger-specific lateral corrections.

This baseline guarantees correct key contact but lacks temporal dynamics.

\subsubsection{Stage 2: Cascaded Trajectory Refinement}
\label{sec:stage2}

Natural piano playing exhibits anticipatory motion toward upcoming keys, smooth follow-through after releases, and context-dependent variations.
We train refinement networks predicting residuals from the baseline, adding these dynamics while preserving geometric constraints through three sub-stages: fingering-conditioned refinement (Stage~2.1), FiLM-based MIDI-conditioned refinement (Stage~2.2), and SmoothNet-style~\cite{zeng2022smoothnet} temporal smoothing (Stage~2.3).

\textbf{Stage 2.1--2.2: Cascaded Refinement.}
We refine trajectories through cascaded networks predicting residuals.
The first refinement network conditions on fingering via a Transformer encoder (4 layers, 8 heads):
$\mathbf{p}_\text{refined1} = \mathbf{p}_\text{base} + \mathcal{R}_1(\mathbf{p}_\text{base}, \mathbf{F})$.
The second incorporates musical context via FiLM~\cite{perez2018film} conditioning on Aria~\cite{bradshawscaling} embeddings $\mathbf{h}_\text{midi}$:
\begin{equation}
    \mathbf{p}_\text{refined2} = \mathbf{p}_\text{refined1} + \mathcal{R}_2(\mathbf{p}_\text{refined1}, \mathbf{F}, \mathbf{h}_\text{midi})
\end{equation}

where $\mathbf{h}_\text{midi} \in \mathbb{R}^{T \times 512}$ are MIDI embeddings from the pre-trained Aria model~\cite{bradshawscaling}. Aria was chosen for its state-of-the-art performance on music understanding benchmarks and its explicit modeling of expressive timing and dynamics---features directly relevant to our musical conditioning objective.
FiLM modulates intermediate features:
\begin{equation}
    \text{FiLM}(\mathbf{x}; \boldsymbol{\gamma}, \boldsymbol{\beta}) = \boldsymbol{\gamma} \odot \mathbf{x} + \boldsymbol{\beta}
\end{equation}
where $(\boldsymbol{\gamma}, \boldsymbol{\beta})$ are predicted from MIDI context via a 2-layer MLP.
This enables hand motion adaptation to musical dynamics---for instance, generating larger preparatory arm lifts and more decisive finger attacks before fortissimo chords, while producing gentler, more sustained finger approaches during pianissimo passages.

\textbf{Training Objective.}
Both refinement networks minimize:
\begin{equation}
    \mathcal{L}_\text{refine} = \lambda_\text{pos} \|\mathbf{p}_\text{pred} - \mathbf{p}_\text{gt}\|_2 + \lambda_\text{vel} \|\dot{\mathbf{p}}_\text{pred} - \dot{\mathbf{p}}_\text{gt}\|_2
\end{equation}
The velocity term encourages temporally coherent refinements.
Residuals are clamped to $\pm 80$mm to prevent pathological corrections. During key presses, we apply \textbf{hard geometric masking}: Y-axis and Z-axis residuals are set to zero. In our piano-aligned coordinate system (X: front-back, Y: along keyboard, Z: height), Y determines \textit{which key} is contacted---lateral drift risks adjacent-key errors---and Z determines press depth---vertical drift causes missed contacts. Only X-axis residuals (contact point along key depth) permit learned refinement, as this dimension has minimal impact on key identity.

\textbf{Stage 2.3: Temporal Smoothing.}
We apply SmoothNet-style~\cite{zeng2022smoothnet} filtering:
\begin{equation}
    \mathbf{p}_\text{smooth} = \mathcal{S}(\mathbf{p}_\text{refined2}, \mathbf{m}_\text{press})
\end{equation}
where $\mathbf{m}_\text{press} \in \{0, 1\}^{T \times 5}$ indicates active presses, with hard masking to preserve Y-axis and Z-axis positions during presses.

Left and right hands are processed separately.

\subsubsection{Stage 3: Wrist Trajectory Estimation}
\label{sec:stage3}

Before synthesizing full hand poses, we estimate wrist trajectories that position each hand optimally relative to target keys.

\textbf{Key-Based Offset Prior.}
We compute per-region wrist offsets from training data.
The keyboard is divided into 8 pitch-based regions (keys 0--15, 15--25, 25--35, 35--45, 45--55, 55--65, 65--75, 75--88), and for each region $k$ we compute the mean offset between wrist position and active fingertip centroid:
\begin{equation}
    \boldsymbol{\delta}_k = \mathbb{E}[\mathbf{p}_\text{wrist} - \bar{\mathbf{p}}_\text{fingertips} \mid \text{active keys} \in k]
\end{equation}
This captures how pianists naturally position their wrists relative to the keys being played in different keyboard regions (bass vs. treble, white vs. black key clusters).

\textbf{Base Wrist Generation.}
For each frame, we compute base wrist position:
\begin{equation}
    \mathbf{p}_\text{wrist,base} = \bar{\mathbf{p}}_\text{fingertips,refined} + \boldsymbol{\delta}_{k_\text{active}}
\end{equation}
where $\bar{\mathbf{p}}_\text{fingertips,refined}$ is the centroid of refined fingertip positions and $k_\text{active}$ is the region containing currently active keys.

\textbf{FiLM-Based Refinement.}
We refine wrist trajectories using the same FiLM mechanism:
\begin{equation}
    \mathbf{p}_\text{wrist} = \mathbf{p}_\text{wrist,base} + \mathcal{R}_\text{wrist}(\mathbf{p}_\text{wrist,base}, \mathbf{F}, \mathbf{h}_\text{midi})
\end{equation}
The refinement network (temporal CNN with 6 residual blocks) adds anticipatory positioning and stylistic variation.
Residuals are clamped to $\pm 50$mm, followed by temporal smoothing.

\subsubsection{Stage 4: STGCN-Based Hand Pose Synthesis}
\label{sec:stage4}

Given wrist position $\mathbf{p}_\text{wrist} \in \mathbb{R}^3$ and fingertip positions $\mathbf{p}_\text{tip} \in \mathbb{R}^{5 \times 3}$ from previous stages, we predict intermediate joint positions (MCP, PIP, DIP) using a Spatio-Temporal Graph Convolutional Network~\cite{stgcn}.
Unlike analytical inverse kinematics that produces unnatural configurations when multiple solutions exist, our learning-based approach captures natural joint coupling observed in professional pianists.

\textbf{Hand Graph Representation.}
We represent each hand as graph $\mathcal{G} = (\mathcal{V}, \mathcal{E})$ with $|\mathcal{V}| = 21$ joints.
Edges $\mathcal{E}$ encode the skeletal hierarchy plus inter-finger MCP connections reflecting palm structure.
We use spatial partitioning with hop distance $\leq 1$ for localized anatomical reasoning.

\textbf{Architecture.}
We employ a UNet-style STGCN with channel multipliers $(64, 128, 256)$ and 2 residual blocks per level.
Each block alternates spatial graph convolution and temporal convolution (kernel size 9).
FiLM layers condition the network on fingering at each level.

During STGCN prediction, wrist and fingertip positions serve as \textit{fixed anchors}---only intermediate joints (15 per hand) are predicted.
The subsequent MIDI-conditioned refinement (Eq.~11) and temporal smoothing operate on the full skeleton for temporal coherence, introducing minor adjustments ($<$2mm on average) to all joints; this two-phase design ensures geometric accuracy during pose synthesis while allowing natural motion dynamics in the final output.
For evaluation, we report F1 on Stage~2.2 output (post-refinement, pre-smoothing) to isolate key contact accuracy from downstream smoothing effects.

\textbf{Anatomical Loss Functions.}
We enforce skeletal validity:
\begin{equation}
    \mathcal{L}_\text{pose} = \mathcal{L}_\text{pos} + \lambda_\text{bone} \mathcal{L}_\text{bone} + \lambda_\text{vel} \mathcal{L}_\text{vel} + \lambda_\text{bio} \mathcal{L}_\text{bio}
\end{equation}
where $\mathcal{L}_\text{pos}$ measures position error, $\mathcal{L}_\text{bone}$ penalizes bone length deviation, $\mathcal{L}_\text{vel}$ encourages smooth velocities, and $\mathcal{L}_\text{bio}$ penalizes biomechanically implausible configurations.

\textbf{MIDI-Conditioned Refinement.}
A final STGCN refinement stage conditions on MIDI embeddings to add expressive variation:
\begin{equation}
    \mathbf{J}_\text{final} = \mathbf{J}_\text{pose} + \mathcal{R}_\text{pose}(\mathbf{J}_\text{pose}, \mathbf{F}, \mathbf{h}_\text{midi})
\end{equation}
followed by temporal smoothing with endpoint constraints preserved.

\subsection{Post-Processing}
\label{sec:postprocessing}

For performances exceeding 8 seconds, we process overlapping windows with 4-second stride and stitch using center-weighted segments with Butterworth filtering at boundaries.
For rendering, we fit the parametric MANO model~\cite{mano} to the keypoint sequence via three-stage optimization achieving 3--5mm fitting error.

\section{Experiments}

\subsection{Experimental Setup}
\label{sec:setup}

\textbf{Dataset.}
We evaluated all models on the FürElise dataset~\cite{10.1145/3680528.3687703}, with our method additionally utilizing the fingering annotations introduced in Section~\ref{sec:dataset}.

\textbf{Implementation.}
MIDI embeddings were extracted using pre-trained Aria~\cite{bradshawscaling}.
Left and right hands are trained separately.
Stages were trained sequentially (1$\to$2$\to$3$\to$4), with each stage using outputs from preceding stages as input.
All training was conducted on a single NVIDIA A6000 (48GB) and completed in approximately 6 hours total.

\textbf{Baseline Details.}
We trained S2C~\cite{10.1145/3746027.3755097} on the FürElise dataset, utilizing its official implementation with the hyperparameters reported in the paper.
Since the official code for FürElise~\cite{10.1145/3680528.3687703} is unavailable, we re-implemented its diffusion model using the official EDGE~\cite{edge} implementation, as described in the paper as the diffusion model architecture. However, this re-implementation suffered from mean collapse, failing to generate distinct finger movements. This resulted in negligible accuracy (F1 $<$ 0.01), so we exclude it from quantitative comparison. Nevertheless, we included it in the qualitative expert assessment (Section~\ref{sec:expert_eval}), as we anticipated that comparing its static movement against other model may yield meaningful insights for the expert assessment.
RoboPianist~\cite{robopianist2023} requires fingering input to specify which finger presses each key; we provided the fingering annotations from the \ours{} Dataset introduced in this work.

\textbf{Automatic Fingering Prediction.}
To evaluate robustness without ground-truth fingering, we tested \textbf{\ours{}$_\text{auto}$} using ArGNN~\cite{10.1145/3503161.3548372}, the open-source state-of-the-art model for fingering prediction.
We fine-tuned on \ours{} Dataset (lr=$5 \times 10^{-5}$, label smoothing=0.3) to adapt from score-based to performance-aligned input, improving General Match Rate from 64.3\% to 68.3\%.

\subsection{Evaluation Metrics}

\textbf{Task Accuracy} measures functional correctness via F1 for key press detection.
We extracted key boundaries from the piano mesh geometry to determine which key each fingertip is over (XY plane), then detect presses when fingertip Z crosses empirically calibrated thresholds ($-1.19$mm for white keys, $+10.38$mm for black keys) derived from ground-truth onset analysis against the Yamaha Disklavier's 8mm acoustic trigger depth.
We used \textit{key-only} matching (correct key regardless of finger) for fair comparison with baselines.

\textbf{Position Accuracy} measures trajectory fidelity:
MPJPE (Mean Per-Joint Position Error) across all 21 joints; Fingertip error (Tip) at 5 fingertip joints.
Units are in mm.

\textbf{Motion Dynamics}:
Acceleration Ratio (Accel.) = $\|\ddot{\mathbf{p}}^{\text{pred}}\| / \|\ddot{\mathbf{p}}^{\text{gt}}\|$.
Values near 1.0 indicate ground-truth-like acceleration magnitude; $<$1.0 indicates under-acceleration (sluggish motion); $>$1.0 indicates over-acceleration (jittery motion). This metric captures magnitude only, not motion \textit{quality} such as anticipatory gestures, which is assessed via expert evaluation (Section~\ref{sec:expert_eval}).

\textit{Note}: RoboPianist~\cite{robopianist2023} optimizes for key contact via physics simulation rather than trajectory matching; position metrics are not applicable as it does not aim to reproduce motion capture trajectories.

\vspace{-3pt}
\subsection{Quantitative Results}

\begin{table*}[t]
\vspace{-3pt}
\caption{Quantitative comparison. $\downarrow$: lower is better, $\uparrow$: higher is better. Accel.\ Ratio$^{*}$: ratio of predicted to GT acceleration magnitude; ${\approx}1.0$ matches GT dynamics, ${<}1.0$ lower acceleration, ${>}1.0$ higher acceleration. F1 scores are evaluated on Stage~2.2 output (post-refinement, pre-smoothing) across all stages to ensure consistent key contact measurement; smoothing improves motion quality but slightly shifts press timing. Position and Accel.\ metrics reflect each stage's actual output.}
\label{tab:sota}
\vspace{-3pt}
\centering
\renewcommand\arraystretch{1.1}
\begin{threeparttable}
\resizebox{\textwidth}{!}{
\begin{tabular}{l|ccc|ccc|ccc|ccc|ccc|ccc|ccc}
\Xhline{1pt}
\multirow{3}{*}{\textbf{Method}} & \multicolumn{3}{c|}{\textbf{Task Accuracy}} & \multicolumn{9}{c|}{\textbf{Position (mm)}$\downarrow$} & \multicolumn{9}{c}{\textbf{Accel. Ratio}$^{*}$} \\
\cline{2-22}
 & \multirow{2}{*}{\textbf{Prec.}$\uparrow$} & \multirow{2}{*}{\textbf{Rec.}$\uparrow$} & \multirow{2}{*}{\textbf{F1}$\uparrow$} & \multicolumn{3}{c|}{\textbf{Fingertip}} & \multicolumn{3}{c|}{\textbf{Wrist}} & \multicolumn{3}{c|}{\textbf{Full}} & \multicolumn{3}{c|}{\textbf{Fingertip}} & \multicolumn{3}{c|}{\textbf{Wrist}} & \multicolumn{3}{c}{\textbf{Full}} \\
\cline{5-22}
 & & & & Left & Right & Both & Left & Right & Both & Left & Right & Both & Left & Right & Both & Left & Right & Both & Left & Right & Both \\
\hline
\textit{FürElise Dataset} & .839 & .670 & .739 & 0.0 & 0.0 & 0.0 & 0.0 & 0.0 & 0.0 & 0.0 & 0.0 & 0.0 & 1.00 & 1.00 & 1.00 & 1.00 & 1.00 & 1.00 & 1.00 & 1.00 & 1.00 \\
\hline
S2C$^{a}$ & .100 & .163 & .121 & 74.1 & 64.3 & 69.2 & 65.4 & 58.5 & 61.9 & 69.6 & 59.9 & 64.8 & 0.74 & 0.67 & 0.70 & 0.75 & 0.79 & 0.77 & 0.77 & 0.72 & 0.75 \\
\rowcolor{gray!10} RoboPianist$^{b}$ & .996$^{\dagger}$ & .783$^{\dagger}$ & .877$^{\dagger}$ & — & — & — & — & — & — & — & — & — & — & — & — & — & — & — & — & — & — \\
\hline
\textbf{\ours{} (Stage 1)} & .798 & \textbf{.991} & .882 & 50.0 & 49.7 & 49.8 & — & — & — & — & — & — & 3.06 & 3.45 & 3.79 & — & — & — & — & — & — \\
\textbf{\ours{} (Stage 2)} & \textbf{.843} & .990 & \textbf{.910} & 38.6 & 36.0 & 37.3 & — & — & — & — & — & — & 2.61 & 2.59 & 2.96 & — & — & — & — & — & — \\
\textbf{\ours{} (Stage 3)} & \textbf{.843} & .990 & \textbf{.910} & — & — & — & \textbf{38.1} & \textbf{35.4} & \textbf{36.8} & — & — & — & — & — & — & \textbf{1.05} & \textbf{0.99} & \textbf{1.02} & — & — & — \\
\textbf{\ours{} (Stage 4)} & \textbf{.843} & .990 & \textbf{.910} & \textbf{37.4} & \textbf{33.3} & \textbf{35.3} & \textbf{38.1} & \textbf{35.4} & \textbf{36.8} & \textbf{34.1} & \textbf{31.5} & \textbf{32.8} & 3.01 & 1.41 & 2.21 & \textbf{1.05} & \textbf{0.99} & \textbf{1.02} & 2.28 & \textbf{1.16} & 1.72 \\
\textbf{\ours{}$_\text{auto}$} & .535 & .899 & .661 & 44.6 & 40.5 & 42.5 & 42.6 & 39.2 & 40.9 & 41.3 & 37.6 & 39.4 & \textbf{1.27} & \textbf{0.80} & \textbf{1.03} & 0.84 & 0.78 & 0.81 & \textbf{1.11} & 0.72 & \textbf{0.91} \\
\Xhline{1pt}
\end{tabular}
}
\begin{tablenotes}
\scriptsize
\item[a] \cite{10.1145/3746027.3755097}, [b] \cite{robopianist2023}.
\item[$*$] Acceleration magnitude ratio; does not capture anticipatory motion quality evaluated in Section~\ref{sec:expert_eval}.
\item[$\dagger$] MuJoCo simulation; not directly comparable (see Sec.~\ref{sec:setup}). Included as physics-based reference.
\end{tablenotes}
\end{threeparttable}
\vspace{-8pt}
\end{table*}

Table~\ref{tab:sota} shows Stage 2's refinement improved F1 from 0.882 to 0.910; Stage 3 incorporated wrist estimation (36.8mm error) with near-GT smoothness (1.02); Stage 4 added full-skeleton synthesis (MPJPE 32.8mm). The pre-smoothing F1 of 0.912 (Table~\ref{tab:ablation_design}) illustrates the accuracy-smoothness trade-off.

The FürElise dataset's F1 (0.739) reflects markerless capture limitations, where individual frames contain tracking noise causing fingertips to appear above or beside the intended key. \ours{} exceeds this because our position prior aggregates statistics across $n \geq 10$ observations per (finger, key) combination: the median position effectively filters out per-frame tracking errors, producing more accurate contact positions than any single ground-truth frame. \ours{}$_\text{auto}$ achieved the most GT-like smoothness (0.91) despite lower F1, suggesting fingering errors introduce variations resembling natural dynamics (Section~\ref{sec:expert_eval}).

\textbf{Statistical Significance.}
Across 27 test pieces, \ours{} achieved F1 = $0.910 \pm 0.028$ with 95\% CI $[0.899, 0.920]$; S2C achieved F1 = $0.121 \pm 0.030$ with 95\% CI $[0.109, 0.132]$; MPJPE = $32.8 \pm 9.9$mm with 95\% CI $[29.2, 36.4]$mm.
Paired t-tests confirmed statistically significant improvement ($t=106.8$, $p<.001$).

\textbf{Analysis by Difficulty and Musical Era.}
We analyzed performance stratified by difficulty level (3--9) and musical era (Baroque through Jazz).
\ours{} maintained consistent F1 scores across all difficulty levels (0.88--0.95) and eras (0.88--0.94), demonstrating robustness to both technical complexity and stylistic variation.

\subsection{User Study}
\label{sec:user_study}

We conducted a perceptual user study to validate our key claims: (1)~\ours{} generates motion with superior quality to baselines, and (2) its quality is comparable to the FürElise dataset.

\textbf{Method.}
$N=41$ participants (ages 20--35) evaluated 130 paired video comparisons using two-alternative forced choice (2AFC).
The video samples were curated by a professional pianist to cover diverse techniques across difficulty levels and musical periods.
Four conditions were compared: FürElise Dataset, \ours{}, S2C~\cite{10.1145/3746027.3755097}, and \ours{}$_\text{auto}$ (using predicted fingering).
Participants rated preferences on three dimensions using 7-point Likert scales: 1) Physical Plausibility, 2) Music-Motion Synchronization, and 3) Musical Expressiveness.
Statistical significance was assessed via one-sample $t$-tests with FDR correction; effect sizes are Cohen's $d$.

\begin{figure}[t]
    \centering
    \vspace{-4pt}
    \includegraphics[width=\columnwidth]{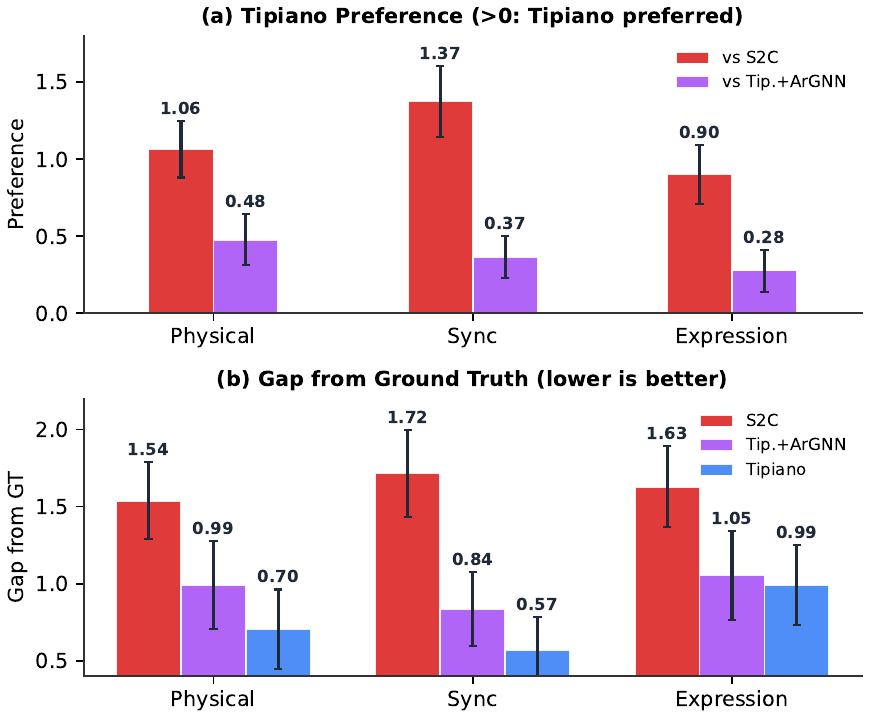}
    \vspace{-8pt}
    \caption{User study results ($N=41$). (a)~Preference for \ours{} over baselines. (b)~Mean preference gap from FürElise dataset on 7-point Likert scale (lower is better; 0 = no preference). \ours{} achieves the smallest gap across all dimensions. Error bars: 95\% CI. All $p<.001$.}
    \label{fig:user_study}
    \vspace{-6pt}
\end{figure}

\textbf{Results.}
Figure~\ref{fig:user_study} summarizes results (all $p<.001$).
\ours{} was strongly preferred over S2C ($d=1.48$--$1.88$).
While FürElise dataset was preferred over \ours{} ($d=0.84$--$1.22$), the perceptual gap was small ($<$1 point on the 7-point Likert scale, corresponding to ``slightly preferred''), indicating \ours{} approaches motion-capture quality.
\ours{} also outperformed \ours{}$_\text{auto}$ ($d=0.64$--$0.91$).
Subgroup analysis showed consistent preferences across expertise levels ($p>.20$).

\subsection{Expert Evaluation}
\label{sec:expert_eval}

We conducted semi-structured interviews with five professional pianists (15--39 years experience) to assess dimensions beyond quantitative metrics.

\textbf{Participants and Procedure.}
Five pianists (15--39 years of performance experience; three with formal teaching experience) participated in 60--90 minute semi-structured interviews.
Experts evaluated FürElise dataset, \ours{}, \ours{}$_\text{auto}$, and S2C on the same excerpts as the user study, with additional comments on FürElise-Diff and RoboPianist.
Transcripts were analyzed using thematic analysis~\cite{braun2006using} following established HCI practices~\cite{mcdonald2019reliability}.

\textbf{Key Findings.}
Four themes emerged from the analysis:

\textit{(1) Over-smoothing eliminates anticipatory motion.}
All experts identified that \ours{} produces trajectories lacking the preparatory movements characteristic of professional playing.
One expert noted: ``\textit{[Tipiano] always returns to a neutral position. With FürElise Dataset, the hand is already preparing for the next note.}''
Another observed that the motion ``\textit{slides smoothly without any sense of actually pressing the keys.}''

\textit{(2) Wrist rigidity undermines perceived naturalness.}
Four experts noted that generated motions appear to originate only from the wrist down, lacking natural vertical movement.
One explained: ``\textit{Even if the coordinates are correct, if the wrist and finger spread look unnatural, it appears wrong.}''

\textit{(3) \ours{}$_\text{auto}$ better conveys musical intent.}
Four of five experts rated \ours{}$_\text{auto}$ as the most ``musical'' among the generated models.
One stated: ``\textit{[\ours{}$_\text{auto}$] looked closest to someone actually playing musically---the thumb crossings and how fingers naturally linger on phrases.}''
This suggests that fingering variation may contribute positively to perceived expressiveness.

\textit{(4) Limited suitability for advanced students.}
Experts agreed that only FürElise dataset motions are appropriate as technique references for conservatory-level students.
Generated models were considered potentially useful for beginners to observe basic positioning, but experts expressed concerns about injury risk: ``\textit{There were moments where I thought, `This looks like it could cause injury.'}''

\textbf{Improvement Priorities.}
Experts recommended incorporating anticipatory and follow-through motion (particularly for thumb-under passages), adding natural wrist vertical movement, and better reflecting attack dynamics for staccato passages.
Experts also commented on additional baselines: FürElise outputs exhibited minimal finger movement and positional drift, consistent with the re-implementation's failure to achieve meaningful contact accuracy; RoboPianist motions were perceived as mechanical rather than human-like.

\subsection{Ablation Study}
\label{sec:ablation}

Table~\ref{tab:ablation_design} validates key design choices in Stage~2.

\begin{table}[t]
\vspace{-6pt}
\caption{Design choice ablation on the test set. All variants are compared at the Stage~2.2 output.}
\label{tab:ablation_design}
\vspace{-3pt}
\centering
\small
\begin{tabular}{l|cccc}
\toprule
Configuration & Prec$\uparrow$ & Recall$\uparrow$ & F1$\uparrow$ & $\Delta$F1 \\
\midrule
Full (proposed) & \textbf{0.847} & \textbf{0.988} & \textbf{0.912} & --- \\
w/o Y-axis masking & 0.846 & 0.909 & 0.876 & $-$0.036 \\
Raw MIDI features & 0.829 & 0.986 & 0.901 & $-$0.011 \\
Concat (no FiLM) & 0.838 & 0.987 & 0.907 & $-$0.005 \\
\bottomrule
\end{tabular}
\end{table}

\textbf{Y-axis Masking} enforces that predicted fingertip positions maintain correct key contact during key presses (Y determines which key; Z determines press depth).
Removing this constraint causes a 8.0\% drop in recall (0.988$\to$0.909), indicating that the model fails to maintain proper key contact without explicit geometric enforcement.
This validates our design of applying hard constraints at the appropriate stage rather than relying on the network to learn them implicitly.

\textbf{Aria MIDI Embedding} provides learned musical representations from a foundation model.
Replacing it with hand-crafted raw MIDI features (452-dimensional pitch, velocity, and timing statistics) reduces F1 by 1.1\%, suggesting that Aria's pre-trained representations capture musically relevant patterns beyond explicit feature engineering.

\textbf{FiLM Conditioning} modulates intermediate features based on MIDI context.
Replacing FiLM with simple concatenation causes a modest 0.5\% F1 drop, indicating that feature-wise modulation provides more effective conditioning than naive concatenation.

\section{Discussion}

\textbf{Resolving the Precision-Naturalness Trade-off.}
Prior work framed precision and naturalness as competing objectives.
Our results demonstrate this trade-off stems from monolithic modeling.
Decomposing synthesis according to the natural constraint hierarchy, \ours{} achieves F1 = 0.910 with Accel.\ Ratio = 1.72, validating that precision and naturalness are complementary when constraints are applied at appropriate abstraction levels. Note that Accel.\ Ratio ${>}1.0$ reflects higher acceleration magnitude, not motion quality. Experts identified insufficient \textit{anticipatory} motion which this metric cannot capture; future metrics should address phrase-level dynamics.

\textbf{Denoising Through Statistical Aggregation.}
\ours{} exceeds its training data accuracy (F1 = 0.910 vs.\ 0.739), demonstrating that establishing geometrically grounded anchors through robust statistics \textit{before} learning dynamics reduces sensitivity to individual frame errors rather than propagating them through end-to-end generation.

\textbf{Limitations.}
First, our method requires fingering annotations at inference.
While \ours{}$_\text{auto}$ partially addresses this (F1 drop to 0.661), expert evaluation suggests degraded output remains suitable for amateur education.
End-to-end joint training remains promising future work.
Second, independent hand processing can produce suboptimal coordination during extreme crossings; explicit bimanual modeling~\cite{10.1145/3746027.3755097} could address this.
Third, our position prior derives from 15 pianists on a single piano model.
While tight prior variance ($\sigma < 17$mm) suggests biomechanical constraints dominate individual differences, populations with different hand anthropometrics may require prior re-estimation.
Our contribution is the \textit{hierarchical framework} itself; automatic prior construction enables adaptation without architectural changes.
Fourth, non-playing gestures (page turns, rest positions) are not modeled, elevating MPJPE; our gesture boundary annotations enable future work on this.

\textbf{Expert Insights.}
Perceived naturalness depends critically on anticipatory motion and wrist dynamics---aspects our smoothing tends to attenuate.
Experts rated \ours{}$_\text{auto}$ as more ``musical'' than \ours{} despite lower F1, while general users prioritized accuracy.
\ours{}$_\text{auto}$'s fingering errors force hand repositioning; experts perceived these larger displacements as preparatory motion, suggesting explicit anticipatory modeling as future work.

\section{Conclusion}

We presented \ours{}, a hierarchical framework exploiting the natural constraint structure of piano motion through a four-stage cascade combining deterministic priors with learned refinement.
By solving the most constrained sub-problem first---fingertip positioning---then progressively adding freedom, \ours{} achieves F1 = 0.910, substantially outperforming diffusion baselines (F1 = 0.121).
User study ($N=41$) confirms quality approaching motion capture, while expert evaluation ($N=5$) identifies anticipatory motion and wrist dynamics as key directions for improvement.

We contribute expert-annotated fingerings for FürElise: 340K per-frame finger-key assignments with gesture boundaries---the first large-scale resource with dense contact supervision.
We release the dataset, code, and trained models to support future research.

The hierarchical decomposition extends beyond piano to any instrument where functional constraints can be factored by ambiguity level, including string instruments and percussion.
More broadly, explicit constraint hierarchies can outperform end-to-end learning for motion synthesis with mixed constraints---a principle applicable to sign language generation, surgical robotics, and industrial manipulation where precision and naturalness must coexist.

\bibliographystyle{ACM-Reference-Format}
\bibliography{references}

@article{NAKAMURA202068,
title = {Statistical learning and estimation of piano fingering},
journal = {Information Sciences},
volume = {517},
pages = {68-85},
year = {2020},
issn = {0020-0255},
doi = {https://doi.org/10.1016/j.ins.2019.12.068},
url = {https://www.sciencedirect.com/science/article/pii/S0020025519311879},
author = {Eita Nakamura and Yasuyuki Saito and Kazuyoshi Yoshii}
}

@inproceedings{10.1145/3503161.3548372,
author = {Ramoneda, Pedro and Jeong, Dasaem and Nakamura, Eita and Serra, Xavier and Miron, Marius},
title = {Automatic Piano Fingering from Partially Annotated Scores using Autoregressive Neural Networks},
year = {2022},
isbn = {9781450392037},
publisher = {Association for Computing Machinery},
address = {New York, NY, USA},
url = {https://doi.org/10.1145/3503161.3548372},
doi = {10.1145/3503161.3548372},
booktitle = {Proceedings of the 30th ACM International Conference on Multimedia},
pages = {6502–6510},
numpages = {9},
location = {Lisboa, Portugal},
series = {MM '22}
}

@inproceedings{kim2025pianovam,
  title={PianoVAM: A Multimodal Piano Performance Dataset},
  author={Kim, Yonghyun and Park, Junhyung and Bae, Joonhyung and Kim, Kirak and Kwon, Taegyun and Lerch, Alexander and Nam, Juhan},
  booktitle={Proceedings of the 26th International Society for Music Information Retrieval Conference (ISMIR)},
  year={2025}
}

@article{mcdonald2019reliability,
  title={Reliability and Inter-rater Reliability in Qualitative Research: Norms and Guidelines for CSCW and HCI Practice},
  author={McDonald, Nora and Schoenebeck, Sarita and Forte, Andrea},
  journal={Proceedings of the ACM on Human-Computer Interaction},
  volume={3},
  number={CSCW},
  articleno={72},
  pages={1--23},
  year={2019},
  publisher={ACM}
}

@article{braun2006using,
  title={Using thematic analysis in psychology},
  author={Braun, Virginia and Clarke, Victoria},
  journal={Qualitative research in psychology},
  volume={3},
  number={2},
  pages={77--101},
  year={2006},
  publisher={Taylor \& Francis}
}

@inproceedings{perez2018film,
  author    = {Perez, Ethan and Strub, Florian and de Vries, Harm and Dumoulin, Vincent and Courville, Aaron},
  title     = {FiLM: Visual Reasoning with a General Conditioning Layer},
  booktitle = {AAAI Conference on Artificial Intelligence},
  year      = {2018}
}

@inproceedings{stgcn,
  author    = {Yan, Sijie and Xiong, Yuanjun and Lin, Dahua},
  title     = {Spatial Temporal Graph Convolutional Networks for Skeleton-Based Action Recognition},
  booktitle = {AAAI Conference on Artificial Intelligence},
  year      = {2018}
}

@article{mano,
  author    = {Romero, Javier and Tzionas, Dimitrios and Black, Michael J.},
  title     = {Embodied Hands: Modeling and Capturing Hands and Bodies Together},
  journal   = {ACM Transactions on Graphics (SIGGRAPH)},
  volume    = {36},
  number    = {6},
  year      = {2017}
}

@inproceedings{bradshawscaling,
  title={Scaling Self-Supervised Representation Learning for Symbolic Piano Performance},
  author={Bradshaw, Louis and Fan, Honglu and Spangher, Alexander and Biderman, Stella and Colton, Simon},
  booktitle={arXiv preprint},
  year={2025},
  url={https://arxiv.org/abs/2506.23869}
}

@inproceedings{zeng2022smoothnet,
  title={Smoothnet: A plug-and-play network for refining human poses in videos},
  author={Zeng, Ailing and Yang, Lei and Ju, Xuan and Li, Jiefeng and Wang, Jianyi and Xu, Qiang},
  booktitle={European Conference on Computer Vision},
  pages={625--642},
  year={2022},
  organization={Springer}
}

@article{zhu2013system,
  title={A system for automatic animation of piano performances},
  author={Zhu, Yuanfeng and Ramakrishnan, Ajay Sundar and Hamann, Bernd and Neff, Michael},
  journal={Computer Animation and Virtual Worlds},
  volume={24},
  number={5},
  pages={445--457},
  year={2013},
  publisher={Wiley Online Library}
}

@inproceedings{10.1145/3680528.3687703,
author = {Wang, Ruocheng and Xu, Pei and Shi, Haochen and Schumann, Elizabeth and Liu, C. Karen},
title = {F\"{u}rElise: Capturing and Physically Synthesizing Hand Motion of Piano Performance},
year = {2024},
isbn = {9798400711312},
publisher = {Association for Computing Machinery},
address = {New York, NY, USA},
url = {https://doi.org/10.1145/3680528.3687703},
doi = {10.1145/3680528.3687703},
booktitle = {SIGGRAPH Asia 2024 Conference Papers},
articleno = {77},
numpages = {11},
location = {Tokyo, Japan},
series = {SA '24}
}

@inproceedings{10.1145/3746027.3755097,
author = {Liu, Zihao and Ou, Mingwen and Xu, Zunnan and Huang, Jiaqi and Han, Haonan and Li, Ronghui and Li, Xiu},
title = {Separate to Collaborate: Dual-Stream Diffusion Model for Coordinated Piano Hand Motion Synthesis},
year = {2025},
isbn = {9798400720352},
publisher = {Association for Computing Machinery},
address = {New York, NY, USA},
url = {https://doi.org/10.1145/3746027.3755097},
doi = {10.1145/3746027.3755097},
booktitle = {Proceedings of the 33rd ACM International Conference on Multimedia},
pages = {9743–9752},
numpages = {10},
location = {Dublin, Ireland},
series = {MM '25}
}

@inproceedings{gan2024pianomotion,
  title={PianoMotion10M: Dataset and Benchmark for Hand Motion Generation in Piano Performance},
  author={Gan, Qijun and Wang, Song and Wu, Shengtao and Zhu, Jianke},
  year={2024},
}

@inproceedings{robopianist2023,
    author = {Zakka, Kevin and Wu, Philipp and Smith, Laura and Gileadi, Nimrod and Howell, Taylor and Peng, Xue Bin and Singh, Sumeet and Tassa, Yuval and Florence, Pete and Zeng, Andy and Abbeel, Pieter},
    title = {RoboPianist: Dexterous Piano Playing with Deep Reinforcement Learning},
    booktitle = {Conference on Robot Learning (CoRL)},
    year = {2023},
}

@misc{qian2024pianomimelearninggeneralistdexterous,
      title={PianoMime: Learning a Generalist, Dexterous Piano Player from Internet Demonstrations}, 
      author={Cheng Qian and Julen Urain and Kevin Zakka and Jan Peters},
      year={2024},
      eprint={2407.18178},
      archivePrefix={arXiv},
      primaryClass={cs.CV},
      url={https://arxiv.org/abs/2407.18178}, 
}

@inproceedings{
huang2025pandora,
title={{PANDORA}: Diffusion Policy Learning for Dexterous Robotics Piano Playing with a Train-only {LLM} Expressiveness Reward},
author={Yanjia Huang and Renjie Li and Zhengzhong Tu},
booktitle={AI for Music Workshop},
year={2025},
url={https://openreview.net/forum?id=lroa1yvUd3}
}

@inproceedings{edge,
  author    = {Tseng, Jonathan and Castellon, Rodrigo and Liu, C. Karen},
  title     = {EDGE: Editable Dance Generation from Music},
  booktitle = {IEEE/CVF Conference on Computer Vision and Pattern Recognition},
  year      = {2023},
  pages     = {448--458}
}

@inproceedings{li2021ai,
  author    = {Li, Ruilong and Yang, Shan and Ross, David A. and Kanazawa, Angjoo},
  title     = {AI Choreographer: Music Conditioned 3D Dance Generation with AIST++},
  booktitle = {IEEE/CVF International Conference on Computer Vision},
  year      = {2021}
}

@article{alexanderson2023listen,
  author    = {Alexanderson, Simon and Nagy, Rajmund and Beskow, Jonas and Henter, Gustav Eje},
  title     = {Listen, Denoise, Action! Audio-Driven Motion Synthesis with Diffusion Models},
  journal   = {ACM Transactions on Graphics (Proceedings of SIGGRAPH)},
  volume    = {42},
  number    = {4},
  pages     = {1--20},
  year      = {2023}
}

@inproceedings{shlizerman2018audio,
  author    = {Shlizerman, Eli and Dery, Lucio and Schoen, Hayden and Kemelmacher-Shlizerman, Ira},
  title     = {Audio to Body Dynamics},
  booktitle = {IEEE/CVF Conference on Computer Vision and Pattern Recognition},
  pages     = {7574--7583},
  year      = {2018}
}

@article{andrychowicz2020learning,
  author    = {Andrychowicz, OpenAI: Marcin and Baker, Bowen and Chociej, Maciek and J{\'o}zefowicz, Rafal and McGrew, Bob and Pachocki, Jakub and Petron, Arthur and Plappert, Matthias and Powell, Glenn and Ray, Alex and others},
  title     = {Learning Dexterous In-Hand Manipulation},
  journal   = {International Journal of Robotics Research},
  volume    = {39},
  number    = {1},
  pages     = {3--20},
  year      = {2020}
}

@article{zhang2021manipnet,
  author    = {Zhang, He and Ye, Yuting and Shiratori, Takaaki and Komura, Taku},
  title     = {ManipNet: Neural Manipulation Synthesis with a Hand-Object Spatial Representation},
  journal   = {ACM Transactions on Graphics (Proceedings of SIGGRAPH)},
  volume    = {40},
  number    = {4},
  pages     = {1--14},
  year      = {2021}
}

@inproceedings{yang2022learning,
  author    = {Yang, Yuting and Wei, Xue and others},
  title     = {Learning Continuous Grasping Function with a Dexterous Hand},
  booktitle = {IEEE International Conference on Robotics and Automation},
  year      = {2022}
}

@article{mordatch2012contact,
  author    = {Mordatch, Igor and Todorov, Emanuel and Popovic, Zoran},
  title     = {Contact-Invariant Optimization for Hand Manipulation},
  journal   = {ACM Transactions on Graphics (SIGGRAPH)},
  year      = {2012}
}

@inproceedings{taheri2020grab,
  author    = {Taheri, Omid and Ghorbani, Nima and Black, Michael J. and Tzionas, Dimitrios},
  title     = {GRAB: A Dataset of Whole-Body Human Grasping of Objects},
  booktitle = {European Conference on Computer Vision},
  pages     = {581--600},
  year      = {2020}
}

@inproceedings{fan2023arctic,
  author    = {Fan, Zicong and Taheri, Omid and Tzionas, Dimitrios and Kocabas, Muhammed and Kaufmann, Manuel and Black, Michael J. and Hilliges, Otmar},
  title     = {ARCTIC: A Dataset for Dexterous Bimanual Hand-Object Manipulation},
  booktitle = {IEEE/CVF Conference on Computer Vision and Pattern Recognition},
  pages     = {12943--12954},
  year      = {2023}
}

@inproceedings{moon2020interhand2,
  author    = {Moon, Gyeongsik and Yu, Shoou-I and Wen, He and Shiratori, Takaaki and Lee, Kyoung Mu},
  title     = {InterHand2.6M: A Dataset and Baseline for 3D Interacting Hand Pose Estimation from a Single RGB Image},
  booktitle = {European Conference on Computer Vision},
  pages     = {548--564},
  year      = {2020}
}

@inproceedings{simon2017hand,
  author    = {Simon, Tomas and Joo, Hanbyul and Matthews, Iain and Sheikh, Yaser},
  title     = {Hand Keypoint Detection in Single Images using Multiview Bootstrapping},
  booktitle = {IEEE/CVF Conference on Computer Vision and Pattern Recognition},
  year      = {2017}
}

@inproceedings{wu2023marker,
  author    = {Wu, Yifan and Liu, Haozhe and Wang, Jingyi},
  title     = {Marker-Based 3D Hand Motion Capture for Piano Performance},
  booktitle = {IEEE Conference on Virtual Reality and 3D User Interfaces},
  pages     = {456--465},
  year      = {2023}
}

@inproceedings{li2018skeleton,
  title={Skeleton Plays Piano: Online Generation of Pianist Body Movements from MIDI Performance.},
  author={Li, Bochen and Maezawa, Akira and Duan, Zhiyao},
  booktitle={ISMIR},
  pages={218--224},
  year={2018}
}

@INPROCEEDINGS{5650193,
  author={Yamamoto, Kazuki and Ueda, Etsuko and Suenaga, Tsuyoshi and Takemura, Kentaro and Takamatsu, Jun and Ogasawara, Tsukasa},
  booktitle={2010 IEEE/RSJ International Conference on Intelligent Robots and Systems}, 
  title={Generating natural hand motion in playing a piano}, 
  year={2010},
  volume={},
  number={},
  pages={3513-3518},
  doi={10.1109/IROS.2010.5650193}}

\end{document}